\documentclass[sigconf]{acmart}
 \pdfoutput=1
\renewcommand\footnotetextcopyrightpermission[1]{} % removes footnote with conference information in first column
\usepackage{balance}
\usepackage{color}
\usepackage[T1]{fontenc}
\usepackage{booktabs}
\usepackage{graphicx, multirow}
\usepackage{placeins}
\usepackage{leftidx}
\usepackage{multirow}
\usepackage{multicol}
\usepackage{array}
\usepackage[hyperindex,breaklinks]{hyperref}
\usepackage{textcomp}
\usepackage{color}
\usepackage{placeins}
\usepackage{caption}
\usepackage{lipsum}
\usepackage{rotating}
\def\BibTeX{{\rm B\kern-.05em{\sc i\kern-.025em b}\kern-.08emT\kern-.1667em\lower.7ex\hbox{E}\kern-.125emX}}
    
\begin{document}

\fancyhead{}
  % do not delete this code.

% The "title" command has an optional parameter, allowing the author to define a "short title" to be used in page headers.
\title{Gradual Network for Single Image De-raining}

% The "author" command and its associated commands are used to define the authors and their affiliations.
% Of note is the shared affiliation of the first two authors, and the "authornote" and "authornotemark" commands
% used to denote shared contribution to the research.

\author{Zhe Huang$^{1\dagger}$, Weijiang Yu$^{2\dagger}$, Wayne Zhang$^{3}$, Litong Feng$^{3}$, Nong Xiao$^{2}$}
\authornote{indicates corresponding author, $\dagger$indicates equal contribution ranked by coin toss. Work done while Zhe Huang \& Weijiang Yu were interns at Sensetime. This is an arXiv preprint, to be presented at ACM Multimedia 2019.}
\affiliation{%
	\institution{$\leftidx^1$University of Wisconsin-Madison, $\leftidx^2$Sun Yat-sen University, $\leftidx^3$SenseTime Research}
}
\email{zhuang334@wisc.edu, weijiangyu8@gmail.com}
\email{{wayne.zhang, fenglitong}@sensetime.com, xiaon6@sysu.edu.cn}

%
% The abstract is a short summary of the work to be presented in the article.
\begin{abstract}
Most advances in single image de-raining meet a key challenge, which is removing rain streaks with different scales and shapes while preserving image details. Existing single image de-raining approaches treat rain-streak removal as a process of pixel-wise regression directly. However, they are lacking in mining the balance between over-de-raining (e.g. removing texture details in rain-free regions) and under-de-raining (e.g. leaving rain streaks). In this paper, we firstly propose a coarse-to-fine network called \textbf{Gradual Network (GraNet)} consisting of coarse stage and fine stage for delving into single image de-raining with different granularities. Specifically, to reveal coarse-grained rain-streak characteristics (e.g. long and thick rain streaks/raindrops), we propose a coarse stage by utilizing local-global spatial dependencies via a local-global sub-network composed of region-aware blocks. Taking the residual result (the coarse de-rained result) between the rainy image sample (i.e. the input data) and the output of coarse stage (i.e. the learnt rain mask) as input, the fine stage continues to de-rain by removing the fine-grained rain streaks (e.g. light rain streaks and water mist) to get a rain-free and well-reconstructed output image via a unified contextual merging sub-network with dense blocks and a merging block. Solid and comprehensive experiments on synthetic and real data demonstrate that our GraNet can significantly outperform the state-of-the-art methods by removing rain streaks with various densities, scales and shapes while keeping the image details of rain-free regions well-preserved.
\end{abstract}

\maketitle

\begin{figure}[ht]
	\centering\includegraphics[width=1.0\linewidth]{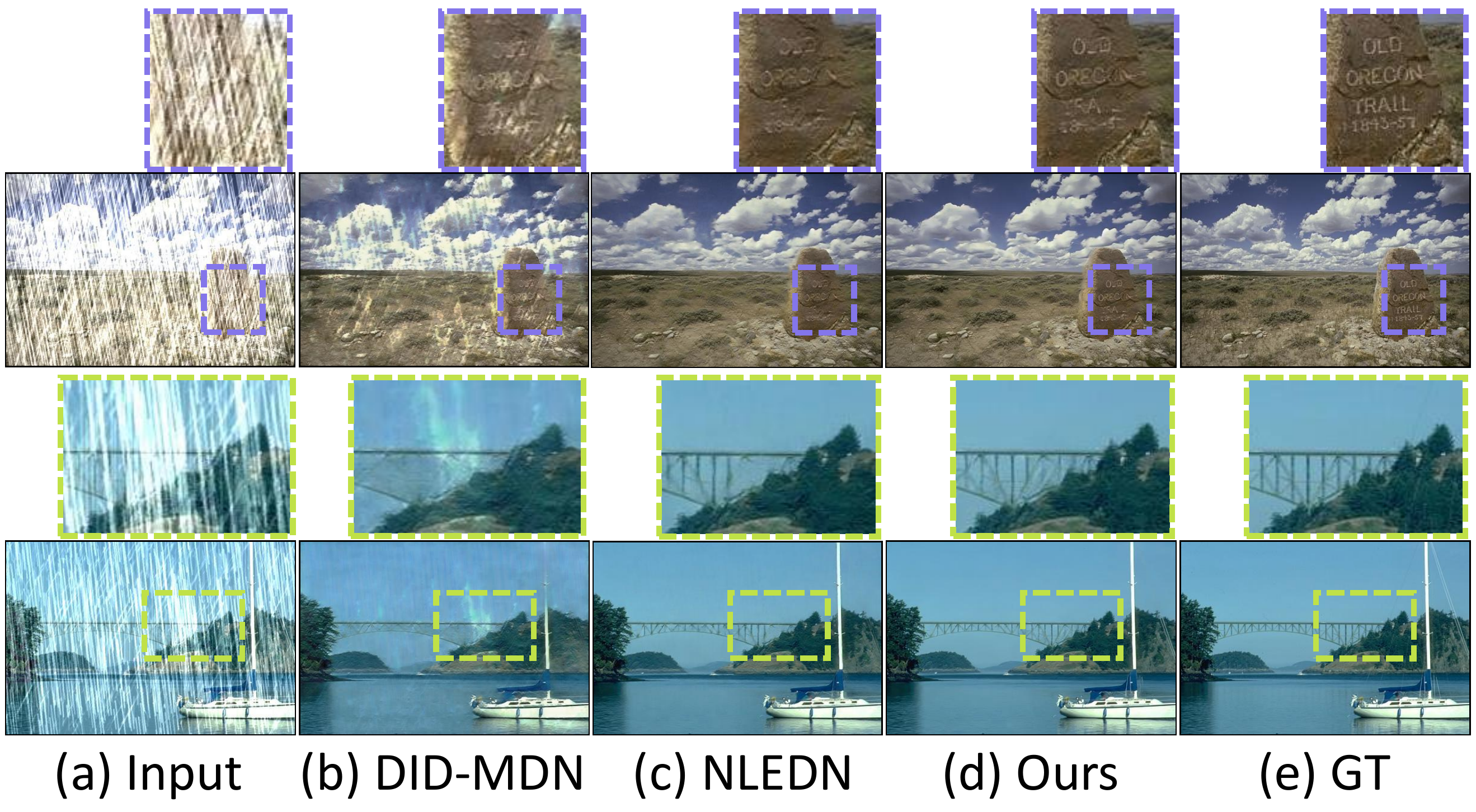}
	%		\vspace{-3mm}
	\caption{Single image de-raining results.  (a) Input rainy images. (b) Results from~\cite{zhang2018density} are under-de-rained with heavy rain streaks. (c) Results from~\cite{li2018non} are over-de-rained with the part of rain-free regions mistakenly removed (e.g. words on the stone, details of the bridge).(d) Our (GraNet) results. (e) Ground truth. Please zoom in the colored PDF version of this paper for more details.}
	\label{fig:pic1}
	%		\vspace{-6mm}
\end{figure}
\section{Introduction}
Diverse weather such as heavy rain, snow or fog will result in complex visual effects on spatial or temporal domains in images or videos~\cite{garg2004detection,garg2005does,garg2006photorealistic}. The impact of rain streaks on images and videos is undesirable due to the visibility degradation, which causes pool performance of many computer vision systems and is accountable for undesirable visual disturbance for some multimedia applications~\cite{xue2012motion,halimeh2009raindrop}. Hence, it is important and necessary to develop algorithms that can remove rain streaks automatically and efficiently. 

In recent years, research on visual rain-streak removal has been gaining increasing attention. Early research mainly focuses on rainy video restoration by de-raining the rain streaks in video sequences~\cite{garg2004detection,kim2015video,zhang2006rain}. They remove rainy streaks mostly by exploiting the temporal correlation between video frames. Some researchers focus on single image de-raining that is more challenging than video de-raining due to the lack of temporal information. 

The problem of single image rain-streak removal is often regarded as a signal separation task by exploring the prior information on physical characteristics of diverse types of rain streaks~\cite{chen2013generalized,sun2014exploiting,fu2017removing}, or an image filtering problem which can be dealt with by restoring the image using non-local mean smoothing~\cite{kim2013single,li2018non}. Fu et al.~\cite{fu2017removing} presented a rain detection and removal network via a recurrent way of modeling physical characteristics of the rain and remove rain streaks. Zhang et al.~\cite{zhang2018density} introduced a network with multi-stream dense connection to model multi-scale features of rain streaks. Based on the analysis of rotation angle and the aspect ratio, Kim et al.~\cite{kim2013single} utilized neighbor pixel correlation via adaptive weight calculation to detect rain streaks. Li et al.~\cite{li2018non} proposed an encoder-decoder network composed of dense non-local operations to solve the problem of single image de-raining. These methods either are aimed at treating rain-streak removal as a smooth process with an end-to-end method, or utilize handcrafted features with strong priori assumptions, modeling rain streaks with specific shapes, scales and density levels. All of them are lacking in effectively and explicitly mining the balance between over-de-raining (e.g. removal of texture details in rain-free regions) and under-de-raining (e.g. leaving rain streaks), which easily gives rises to three limitations: 1) they tend to over-de-rain image to remove some important parts of rain-free regions (Figure~\ref{fig:pic1}(c)). 2) they tend to under-de-rain image to leave some rain streaks especially in heavy rainy scenes (Figure~\ref{fig:pic1}(b)). 3) implicit learning tends to block the interpretability of the network's behavior, causing confusion when trying to understand the de-raining procedure.

\begin{figure*}[t]
	\centering\includegraphics[width=0.9\linewidth]{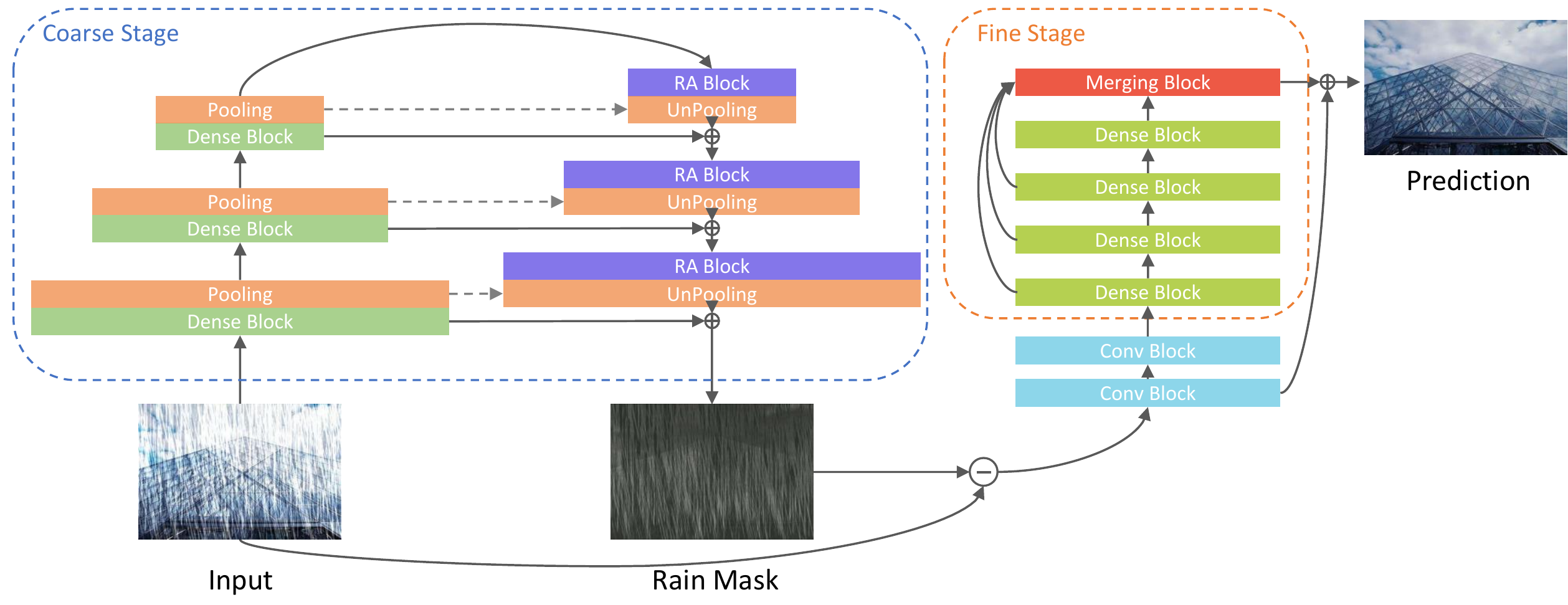}
	\vspace{-3mm}
	\caption{An overview of our Gradual Network (GraNet). Our GraNet consists of coarse stage and fine stage. The coarse stage is aimed to mine coarse-grained rain streaks, which generates local information via dense blocks with pooling mechanism and generates global information via Region-aware (RA) blocks with unpooling mechanism. Those dash lines indicate that unpooling operations are guided under corresponding pooling indices~\cite{badrinarayanan2015segnet} to maintain spatial coordinate information and "$\bigoplus$" denotes the fusion of local and global information. The fine stage is used to remove fine-grained rain streaks from coarse de-rained result to get the final rain-free image via a unified contextual merging sub-network composed of dense blocks and a merging block.}
	\label{fig:framework}
	%	\vspace{-6mm}
\end{figure*}

To address all above-mentioned issues, we propose a new model to endow the deep network with the capability of coarse-to-fine de-raining to make rain-streak removal procedure coherent and gradual from coarse to fine. We define a coarse-to-fine network called Gradual Network (GraNet) consisting of coarse stage and fine stage for single image de-raining with different granularity. The GraNet is shown in Figure~\ref{fig:framework}. It effectively detects coarse-grained rain-streak characteristics (e.g. long and thick rain streaks/raindrops) via a coarse stage, and effectively removes fine-grained rain streaks (e.g. light rain streaks and water mist)  progressively via a fine stage. Specifically, the coarse stage is to produce a coarse-grained mask map of rain streaks. Given the result of the coarse stage, a residual subtraction is applied with the input rainy image to explicitly obtain a coarse de-rained result. To further obtain a fine de-raining result, a fine stage is attached to coarse stage to take the coarse de-rained result as input. The fine stage is able to explore the unified contextual information to remove fine-grained rain streaks from coarse de-rained result and produce a satisfactory output via a series of dense blocks and a merging block.

The non-uniform spatial distribution and diverse shapes, scales and densities of rain streaks have reflected the trait that rain steaks has different characteristics locally and globally. Unlike some existing works~\cite{li2018non,kim2013single} which only propagate the visual features of rain streaks globally from all the layers to refine the final regression result directly, we first consider the local-global joint information to extend well beyond the region of interest and mitigate the influence of irrelevant image content. We define the coarse stage as a local-global sub-network to focus on modeling coarse-grained rain streaks. To achieve this, at first, our coarse stage generates local information made up of shallow features via specific convolutional filters (e.g. $3\times 3$ convolution kernel with dense connection) with pooling mechanism. Note that densely-connected convolutional structure at the very beginning can effectively extract low-level local features and reduce the overuse of convolution layers. Then given the local features, our coarse stage produces global information via region-aware blocks composed of non-local operations~\cite{NonLocal2018} over all deep features. Finally, the local information and global information are fused by guided pooling indices in the coarse stage to improve its feature representation on pertinent rainy regions.

Given the rain streak mask produced by the coarse stage, we explicitly remove the coarse-grained rain streaks by a residual subtraction to obtain coarse de-rained result. Based on the coarse de-rained result with coarse-grained rain streaks already being removed, we use a fine stage as a unified contextual merging sub-network to remove fine-grained rain streaks while preserving rain-free details. Specifically, our fine stage generates features with various receptive fields from low-level features to high-level features via a series of dense blocks, and then accumulates the generated multi-level features via a merging block to obtain aggregated representations with stronger capability for fine-grained rain streaks removal while maintaining and reinforcing image details.      

To further demonstrate the effectiveness of GraNet, we evaluate the GraNet on four de-raining benchmarks (e.g. DDN-Dataset~\cite{fu2017removing}, DID-Dataset~\cite{zhang2018density}, Rain100L~\cite{yang2017deep}, Rain100H~\cite{yang2017deep}) compared with state-of-the-art methods to show its superior performance. And we also test our GraNet on real-world rainy images to demonstrate its generalization ability and versatility. 

Our contributions are summarized into the following aspects. 

\textbf{1)} We firstly propose a coarse-to-fine network called Gradual Network (GraNet) for single image de-raining with different granularities, which consists of coarse stage and fine stage for mining the balance between over-de-raining and under-de-raining. To the best of our knowledge, it is the first strategy dividing the de-raining task into coarse de-raining and fine de-raining.

\textbf{2)} We consider the local-global joint feature representation to extend well beyond the region of interest and mitigate the influence of irrelevant image content, and define a coarse de-raining stage to mine coarse-grained rain streaks via a local-global sub-network contained region-aware blocks. 

\textbf{3)} We define a fine de-raining stage as a unified contextual merging sub-network to remove fine-grained rain streaks while preserving rain-free region details.

\textbf{4)} Solid and comprehensive experiments on synthetic datasets and real data demonstrate that our GraNet can significantly outperform the state-of-the-art methods by removing rain streaks with various densities and shapes while well-preserving the image details of rain-free regions.

\section{Related Work}
Due to the powerful feature representation by utilizing deep convolutional neural networks, the research on pixel-wise mapping of the image has become a key area in the field of computer vision and multimedia, such as image and video super-resolution~\cite{shi2016real,kim2016deeply,mao2016image} and de-raining~\cite{santhaseelan2015utilizing,luo2015removing,wang2019spatial}.
The de-raining domain has been divided into two categories: video-based de-raining task and single image based de-raining task. For video-based task, researchers often use inter-frame information to remove the rain streaks~\cite{chen2013generalized,zhang2006rain,roser2009video}. For single image based de-raining task, it is more ill-posed due to the lack of temporal information for detecting and removing rain streaks. Most advances in single image de-raining meet a key challenge of removing rain streaks with different scales and shapes while preserving the image details~\cite{garg2005does,kang2012automatic}. 

These methods either are aimed to treat rain-streak removal as a smooth process with an end-to-end method, or utilize handcraft feature representation with strong prior assumptions. For prior-based methods, Kang et al.~\cite{kang2012automatic} presented sparse coding-based method to remove rain streaks. Li et al.~\cite{li2016rain} used layer priors via GMM for rain-streak removal. Based on the analysis of rotation angle and the aspect ratio, Kim et al.~\cite{kim2013single} utilized neighbor pixel correlation via adaptive weight calculation to detect rain streaks. All prior-based methods are limited due to the over-de-raining as they smooth some texture details of rain-free regions.

With the advances in deep learning architectures~\cite{he2015deep,huang2017densely}, recent research has obtained inspiring results via deep convolutional neural networks. For example, Fu et al.~\cite{fu2017removing} presented a rain detection and removal network via a recurrent way of modeling physical characteristics of the rain and remove rain streaks. Zhang et al.~\cite{zhang2018density} introduced a network with multi-stream dense connection to model multi-scale features of rain streaks. Li et al.~\cite{li2018non} proposed an encoder-decoder network composed of dense non-local operations to solve the problem of single image de-raining. All of them lack mining the balance between over-de-raining (e.g. removing texture details in rain-free regions) and under-de-raining (e.g. leaving rain streaks). Hence, we firstly present a coarse-to-fine network for mining single image de-raining with different granularities.

\begin{figure}[t]
	\centering\includegraphics[width=0.9\linewidth]{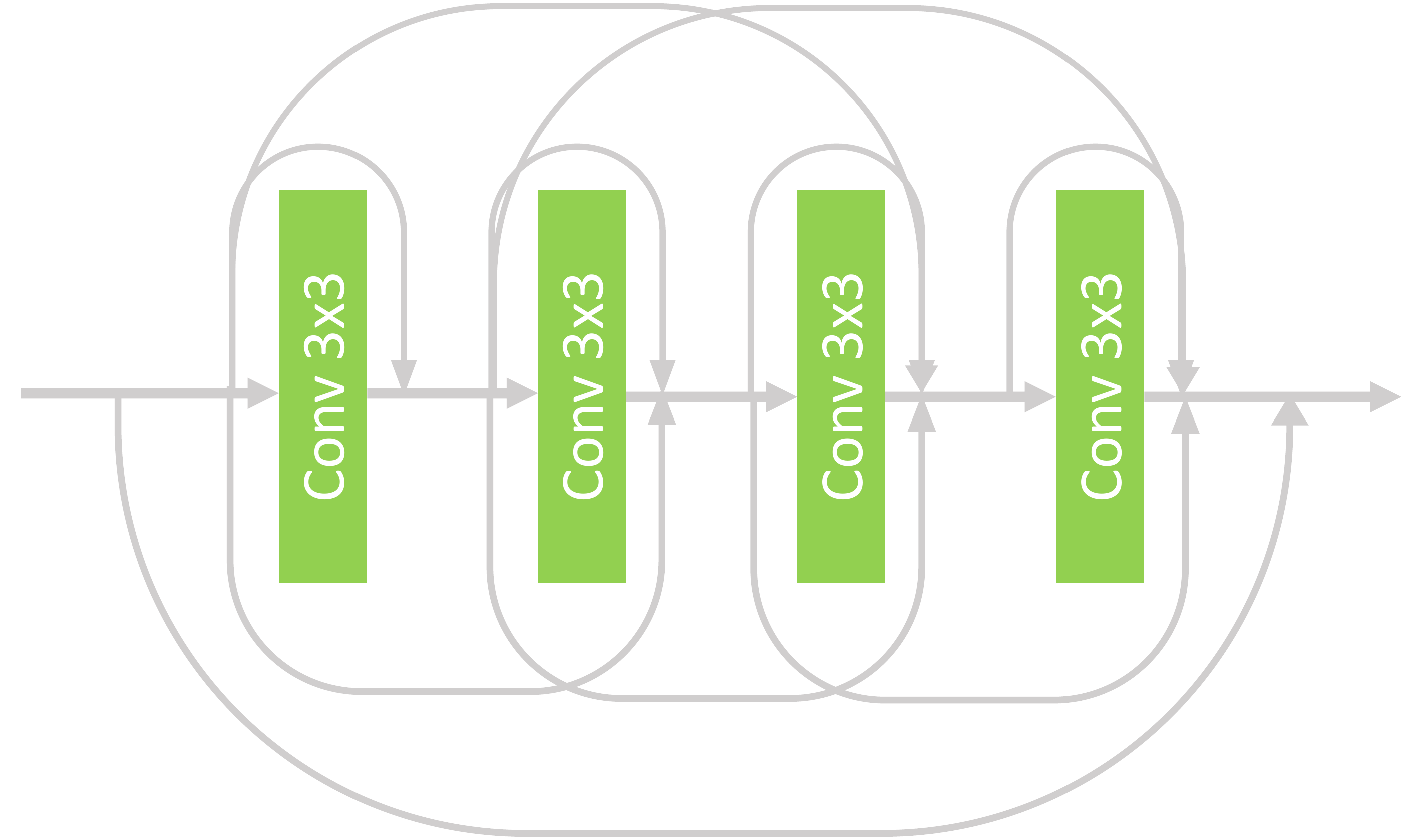}
	\vspace{-3mm}
	\caption{The architecture of the proposed dense block.}
	\label{fig:dense}
	%	\vspace{-6mm}
\end{figure}
\section{Gradual Network}
\subsection{Overview}
In this section, we explain the architecture and design of our proposed Gradual Network (GraNet), which can appropriately remove rain streaks and preserve rain-free image details. Our GraNet consists of two stages: coarse de-raining stage and fine de-raining stage. Given the input rainy image, the proposed coarse stage utilizes convolutional blocks with pooling mechanism to generate multi-scale local feature representations. Taking the local features as input, an inverted feature pyramid is applied for learning the global spatial correlation of each positions via region-aware blocks attached to unpooling mechanism. After fusing the local and global information via guided pooling indices~\cite{badrinarayanan2015segnet} with skip connections, a coarse-grained rain streak mask is produced. The mask map often contains long and thick rain streaks. With the subtraction operation between input rainy image and the produced rain mask, a coarse de-rained result is generated explicitly. Finally, a carefully-designed fine stage that consists of dense blocks and a merging block is utilized to remove fine-grained rain streaks from the coarse de-rained result to get the final rain-free image. The overview of the our GraNet is illustrated in Fig.~\ref{fig:framework}. Whereas the dense block in Fig.~\ref{fig:framework}, which is adopted widely across our two stages, has a simple and easy-to-understand structure as illustrated in Fig.~\ref{fig:dense}, our contribution mainly focuses on the design of a two-stage framework and innovative modules (e.g. Region-aware (RA) Block and Merging Block), which are unveiled and discussed in details in the following sub-sessions.
\subsection{Coarse Stage}
This stage is used to produce coarse-grained rain streaks via a local feature extraction and a global feature extraction. The local feature extraction is made up of a series of residual convolutional blocks~\cite{he2015deep} with maxpooling operations with $2\times 2$ strides. The global feature extraction consists of a series of region-aware blocks with maxunpooling operations that are guided by pooling indices~\cite{badrinarayanan2015segnet} to effectively recover feature maps. 

Firstly, we define the given input rainy image as $I$ and define the $i_{th}$ layer of the local feature extraction as $f_{L}^{(i)}$ that contains convolutional operation and activation function (e.g. ReLU). We have the following formulae
\begin{equation}
F_{L}^{(0)} = f_{L}^{(0)}(I) \quad,
\end{equation}
\begin{equation}
F_{pool}^{(0)} = \psi_{pool}^{(0)}(F_{L}^{(0)}) \quad,
\end{equation}
where $F_{L}^{(0)}$ denotes the first output feature maps of the first residual convolution layer in local feature extraction, and $F_{pool}^{(0)}$ means the first output pooling feature maps of the first maxpool operation. The $\psi_{pool}^{(0)}$ denotes the $0_{th}$ maxpool operation with $2\times 2$ strides, respectively. As shown in Fig.~\ref{fig:framework}, there are three layers in local feature extraction that contains six outputs, denoted as $F_{L}^{(0)}$, $F_{L}^{(1)}$, $F_{L}^{(2)}$, $F_{pool}^{(0)}$, $F_{pool}^{(1)}$ and $F_{pool}^{(2)}$, respectively. We have the formula
\begin{equation}
F_{pool}^{(i+1)} = \psi_{pool}^{(i+1)}(f_{L}^{(i+1)}(F_{L}^{(i)})) \, , \, 0\leq i < N\quad,
\end{equation}
where $i$ means the $i_{th}$ input feature maps to feed into the ${i+1}_{th}$ layer in local feature extraction, $N$ means the number of layers in the whole local feature extraction.
Being similar with local feature extraction, we formulate the global feature extraction as
\begin{equation}
F_{unpool}^{(j)} = \psi_{unpool}^{(j)}(f_{G}^{(j)}(F_{G}^{(j-1)})) + F_{L}^{(i)} \, , \, i+j=N\quad,
\end{equation}
where $F_{unpool}^{(j)}$ denotes the $j_{th}$ ouput unpooling feature maps, $\psi_{unpool}^{(j)}$ means $j_{th}$ unpool operation and $F_{G}^{(j-1)}$ indicates the ${(j-1)}_{th}$ input feature maps. The $f_{G}^{(j)}$ means the $j_{th}$ region-aware block to model global correlation of each position at the feature maps. 

Here we introduce our region-aware block, which is a regional non-local design aimed at enlarging receptive field for better feature representation, and then we will illustrate the implementation of our proposed coarse stage.

\textbf{Region-aware Block} We adopt and modify non-local design introduced by~\cite{NonLocal2018}. The idea is to enlarge the receptive field and to send and gather non-local or global information between intermediate layers. The original representation of non-local structure is formulated as
\begin{equation}
\mathbf{y}_{i}=\frac{1}{\mathcal{C}(\mathbf{x})} \sum_{\forall j} f\left(\mathbf{x}_{i}, \mathbf{x}_{j}\right) g\left(\mathbf{x}_{j}\right)\quad,
\end{equation}
%$$\mathbf{y}_{i}=\frac{1}{\mathcal{C}(\mathbf{x})} \sum_{\forall j} f\left(\mathbf{x}_{i}, \mathbf{x}_{j}\right) g\left(\mathbf{x}_{j}\right)$$ 
where $\mathbf{y}_{i}$, $\mathbf{x}_{i}$ denote the output and the input at position $i$, respectively, and $\mathbf{x}_{j}$ denote the input at position $j$. $\mathcal{C}(\mathbf{x})$ is a normalization factor, which is defined as $\mathcal{C}(\mathbf{x})= \sum_{\forall j}f(\mathbf{x}_{i},\mathbf{x}_{j})$. Using this generic non-local operation, each output collects all input information from every input position. However, in single image de-raining problem, due to the large size of input image and intermediate features, we propose to replace non-local structure with regional non-local structure which is region-aware to keep feasible computation cost. The region-aware non-local operation is defined as
\begin{equation}
\mathbf{y}_{i}^{r}=\frac{1}{\mathcal{C}(\mathbf{x}^{r})} \sum_{j \in r} f\left(\mathbf{x}_{i}^{r}, \mathbf{x}_{j}^{r}\right) g\left(\mathbf{x}_{j}^{r}\right) \quad,
\end{equation}
%$$\mathbf{y}_{i}^{r}=\frac{1}{\mathcal{C}(\mathbf{x}^{r})} \sum_{j \in r} f\left(\mathbf{x}_{i}^{r}, \mathbf{x}_{j}^{r}\right) g\left(\mathbf{x}_{j}^{r}\right)$$ 
where $\mathbf{y}_{i}^{r}$, $\mathbf{x}_{i}^{r}$ denote the output and the input at position $i$ within the region $r$, respectively, and $\mathbf{x}_{j}^{r}$ denote the input at position $j$ within the region $r$.

In our proposed method, since our network is fully convolutional, which means all features for an input sample is 3D shaped such as $C \times H \times W$, where $C$, $H$ and $W$ stand for the number of channels, the height and the width of this feature map, respectively. Thus, the region $r$ here refers to a sub area as $C \times H_{r} \times W_{r}$, where $H_{r}$ and $W_{r}$ refer to the height and the width of this region, which is across all channels but with spatial dimensions reduced.

Different from~\cite{li2018non} that combines non-local operation with dense convolution layers as a fixed and consolidated module applied to the whole network, our region-aware block is more flexible to be incorporated in the network with lighter computation. Thanks to the RA block, our network can flexibly fuse local and global information.

\textbf{Coarse Stage Implementation.} We build our proposed coarse stage as illustrated in Fig.~\ref{fig:framework}. We keep maxpooling indices for corresponding maxunpooling. Empirically, we set the number of regions to be $1\times 1$, $2\times 2$, and $4\times 4$ for three different region-aware modules respectively, meaning that we divide the feature map before region-aware mechanism into $1\times 1$, $2\times 2$, and $4\times 4$ blocks correspondingly for the three RA blocks and then perform our non-local operation on each block for region-aware purpose.

Note that a residual subtraction is utilized between the input rainy image and the coarse-grained rain-streak mask produced by coarse stage to explicitly accomplish coarse de-raining, whose interpretability can be illustrated through our extensive experiments.
\subsection{Fine Stage}
The fine stage is designed to remove fine-grained rain streaks from coarse de-rained result to get the final rain-free image via a unified contextual merging sub-network composed of dense blocks~\cite{huang2017densely} and a merging block. Specifically, we utilize a stacked architecture contains several dense blocks, then merging the output feature maps of different dense blocks by skip connections and a merging block. 
Here we focus on introducing our merging block at fine stage, which is designed to use contextual information for image reconstruction, and then we will introduce the detailed implementation of the proposed fine stage.

\textbf{Merging Block} We design a merging block to perform the fine de-raining to produce the final constructed rain-free image of our proposed GraNet. The functionality of this block can be denoted as
\begin{equation}
B = F_{merge}\left(A\right)\quad,
\end{equation}
%$$B = F_{merge}\left(A\right)$$
where $B$ and $A$ are output and input features respectively. Specifically, we utilize feature maps from different levels to mine contextual information. In this way, features from different levels can adaptively vote to form the final feature representations. The size of the input $A$ is $C \times H \times W$ and the size of the output $B$ is $C^\prime \times H \times W$, where $C^\prime = C / k$. $k$ is the scale factor denotes that how many channels in the input feature $A$ can formulate one channel in the output feature $B$. Inspired by the pixel shuffle operation in~\cite{shi2016real}, our merging operation first separates input feature by channel to form groups as $\left[ A^0, A^1 ... A^{k} \right]$ with the size of each grouped feature to be exactly $C^\prime \times H \times W$. After the operation that we can merge contextual information between low-level and high-level to generate the final rain-free image $B$. Consequently, each output position $B_{c, h, w}$ is calculated as
\begin{equation}
B_{c, h, w} = \frac{1}{k}\sum_{i=1}^{k}A_{c, h, w}^{i}\quad,
\end{equation}
thus the voted output result $B$ is generated by utilizing unified contextual information.

\textbf{Fine Stage Implementation} We build our proposed fine stage in the way illustrated in Fig. ~\ref{fig:framework}. Empirically, we set $k = 4$ in our experiments, meaning that we merge every 4 channels from the input of the merging block into one channel in its output.

\subsection{Loss Function}
We use the mean value of the per-pixel absolute distance between the de-rained image and the ground truth to define our Mean Absolute Error (MAE) loss function, denoted as $\mathcal{L}$, explained as
\begin{equation}
\mathcal{L} = \frac{1}{HWC}\sum_{c}^{}\sum_{w}^{}\sum_{h}^{} \lVert \Phi\left(I\right)_{c,h,w} - Y_{c,h,w} \rVert_1\quad,
\end{equation}
where $\Phi$, $I$ and $Y$ denote the GraNet, the input rainy image and the corresponding rain-free image. The $H$,$W$ and $C$ indicate the height, width and channel number of rainy image.

\section{Experiments}
In this session, we first introduce the data that we use and details of our experiments, and then we demonstrate our results in comparison with other methods followed by our ablation study and the interpretation of our model.

\subsection{Studied Datasets}
We evaluate our proposed de-raining method and compare it with other state-of-the-art methods on four synthetic datasets, which are widely used and accepted for academic de-raining, including Rain100L and Rain100H datasets provided by~\cite{yang2017deep}, the dataset released by Fu et al.~\cite{fu2017removing}, referred as DDN-Dataset, and the dataset created by Zhang et al.~\cite{zhang2018density}, denoted as DID-Dataset here. For details, Rain100L and Rain100H contain rain-streak masked images originated from BSD200 dataset~\cite{martin2001database}. Specifically, the Rain100L dataset only consists of 200 training pairs and 100 testing pairs of synthesized raining images with single style low-density rain streaks, whereas the Rain100H dataset contains 1800 training pairs with five different types of heavy rain-streak masks and another 100 pairs for testing. The DDN-Dataset created by Fu et al.~\cite{fu2017removing}, is made up of 14,000 image pairs via synthesizing 1000 clean images with 14 different artificial rain-streak masks. To be consistent with Li et al.~\cite{li2018non}, we split this dataset into 9,100 training images pairs and 4,900 testing pairs. The DID-Dataset generated by Zhang et al.~\cite{zhang2018density} has 12,000 image pairs which can be divided into three categories based on the density of rain streaks masked on each raining image, from light to heavy. For each training pair, a label indicating the density is also available, which is utilized as density information by their~\cite{zhang2018density} density-aware de-raining strategy. We would like to emphasize that neither does our method require nor do we use this extra information for better performance.

\subsection{Evaluation Criteria}
As other works do~\cite{li2018non,zhang2018density,yang2017deep,fu2017removing}, for four synthetic datasets, we also measure the performance of our proposed method via two metrics, which are the Peak Signal-to-Noise Ratio (PSNR)~\cite{huynh2008scope},  and the Structure Similarity Index (SSIM)~\cite{wang2004image}. We convert our predicted results into $YC_bC_r$ color space and calculate those indices on the luminance channel (i.e. $Y$ channel) of our results. For real-world performance, since there are no ground truth for real-world data, we evaluate them visually.
\begin{table*}[]
	\tabcolsep 0.06in{\scriptsize{}}
	\begin{tabular}{c|c|c|c|c|c|c|c|c}
		\hline
		Dataset & Metric & \begin{tabular}[c]{@{}c@{}}DSC {~\cite{luo2015removing}}\\ (ICCV '15)\end{tabular} &  
		\begin{tabular}[c]{@{}c@{}}GMM {~\cite{li2016rain}}\\ (CVPR '16)\end{tabular} & 
		\begin{tabular}[c]{@{}c@{}}DDN {~\cite{fu2017removing}}\\ (CVPR '17)\end{tabular} & 
		\begin{tabular}[c]{@{}c@{}}JORDER {~\cite{yang2017deep}}\\ (CVPR '17)\end{tabular} & 
		\begin{tabular}[c]{@{}c@{}}DID-MDN {~\cite{zhang2018density}}\\ (CVPR '18)\end{tabular} & 
		\begin{tabular}[c]{@{}c@{}}NLEDN {~\cite{li2018non}}\\ (MM '18)\end{tabular} & Our GraNet \\ \hline
		\multirow{2}{*}{DDN-Dataset} & PSNR   & 22.03 & 25.64 & 28.24 & 28.72 & 26.17 & 29.79 & \color{red}{\textbf{32.51}} \\ \cline{2-9} 
		& SSIM   & 0.7985 & 0.8360 & 0.8654 & 0.8740 & 0.8409 & 0.8976 & \color{red}{\textbf{0.9292}} \\ \hline
		\multirow{2}{*}{DID-Dataset} & PSNR   & 20.89 & 21.37 & 23.53 & 30.35 & *28.30 & 33.16 & \color{red}{\textbf{33.68}} \\ \cline{2-9} 
		& SSIM   & 0.7321 & 0.7923 & 0.7057 & 0.8763 & *0.8707 & 0.9192 & \color{red}{\textbf{0.9284}} \\ \hline
		\multirow{2}{*}{Rain100L}    & PSNR   & 23.39 & 28.25 & 25.99 & *35.23 & 30.48 & 36.57 & \color{red}{\textbf{37.55}}\\ \cline{2-9} 
		& SSIM   & 0.8672 & 0.8763 & 0.8141 & *0.9676 & 0.9323 & 0.9747 & \color{red}{\textbf{0.9806}}\\ \hline
		\multirow{2}{*}{Rain100H}    & PSNR   & 17.55 & 15.96 & 23.93 & *26.08 & 26.35 & 30.38 & \color{red}{\textbf{33.16}}\\ \cline{2-9} 
		& SSIM   & 0.5379 & 0.4180 & 0.7430 & *0.8211 & 0.8287 & 0.8939 & \color{red}{\textbf{0.9290}}\\ \hline
	\end{tabular}
	\smallskip
	\noindent\caption{Quantitative comparison w.r.t. to PSNR and SSIM metrics on four synthetic benchmark datasets. The best result of each metric of each dataset is shown in {\color{red}{red}}. As is shown clearly, our proposed GraNet outperforms all other state-of-the-art methods. "*" denotes that additional data (e.g. rain-streak density) is used. Note that we do not need such extra information.}
	\label{tab:comparison}
\end{table*}
\begin{figure*}[t]
	\centering\includegraphics[width=0.9\linewidth]{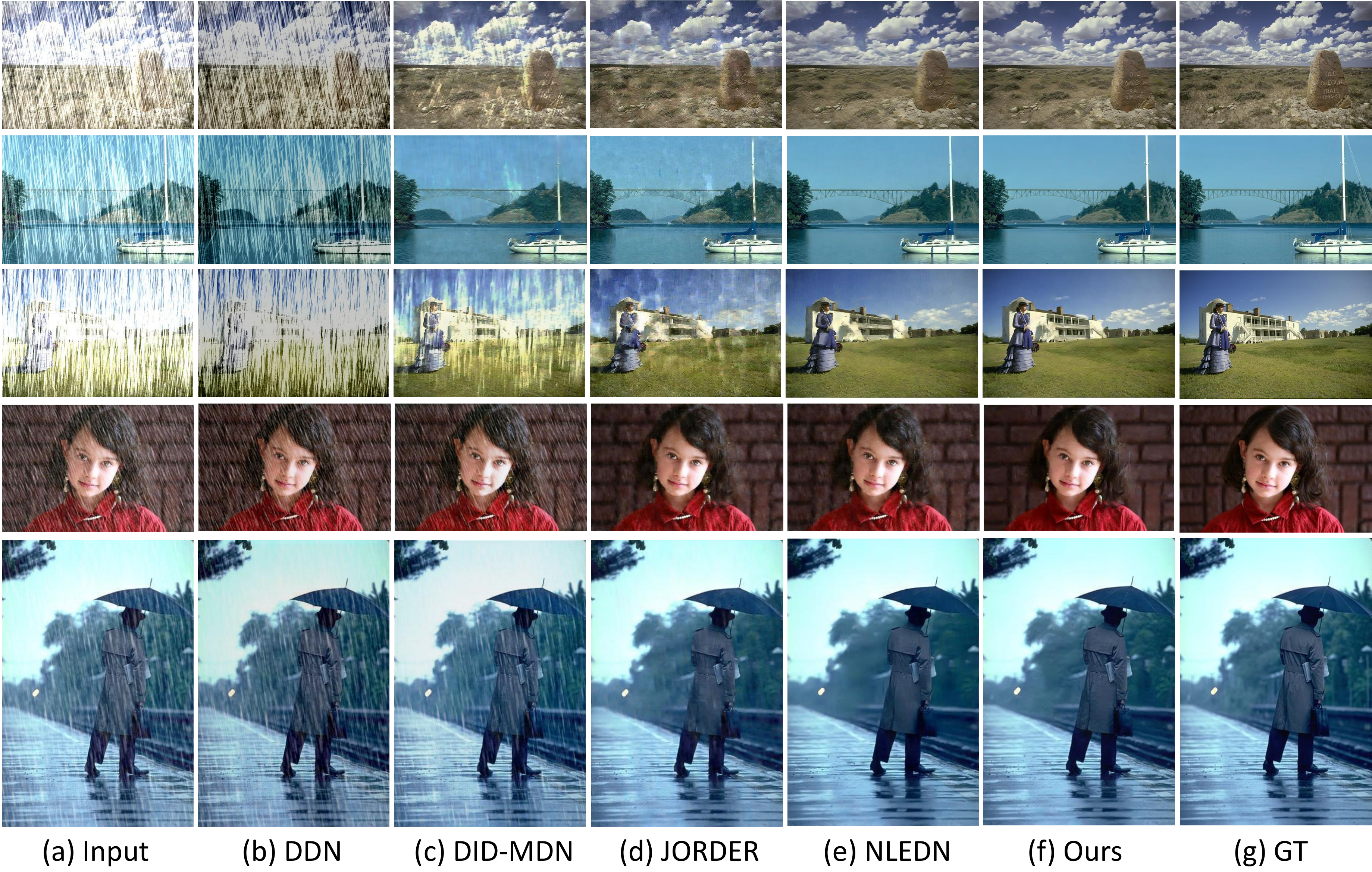}
		\vspace{-3mm}
	\caption{Qualitative visual results on synthetic datasets, which demonstrates the network performance directly via human visual response. Compared with other state-of-the-art methods, our proposed GraNet generates the clearest, cleanest, and the most flawless results. Please zoom in the colored PDF version of this paper for more details.}
	\label{fig:synthetic-qualitative}
		\vspace{-4mm}
\end{figure*}
\begin{figure*}[t]
	\centering\includegraphics[width=0.9\linewidth]{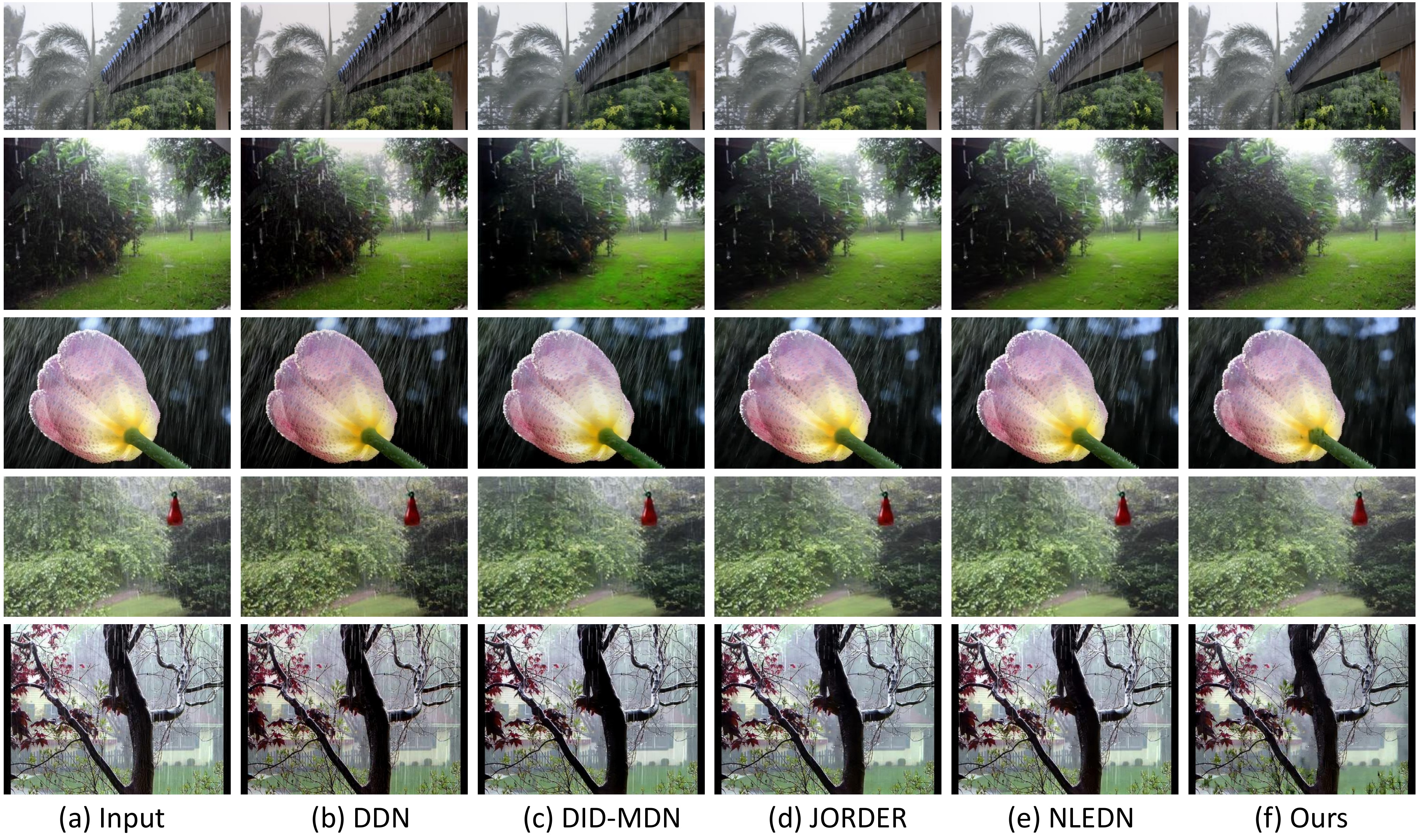}
		\vspace{-3mm}
	\caption{Qualitative results on real-world images, which is another concrete evidence for proving the effectiveness of our proposed method (GraNet). From thin and light rain steaks to thick rain streaks and heavy downpour raining scenes, our model gives out satisfactory results in terms of how it removes rain streaks while keeping original details intact. Please zoom in the colored PDF version of this paper for more details.}
	\label{fig:real-qualitative}
		\vspace{-3mm}
\end{figure*}

\subsection{Implementation Details}
We implement our proposed GraNet via PyTorch v1.0.0~\cite{paszke2017automatic}, a well-known open-source deep learning framework. For being fairly comparable with other methods, we use the following training and testing settings, which are also widely-accepted and away from bells and whistles.

\textbf{Training.} During training, following~\cite{li2018non}, the length of the long side of each input image is constrained to be under or equal to 512 pixels. We utilize the aforementioned MAE loss $\mathcal{L}$ for all experiments. We set the batch size to be 1 per GPU and we use Adam optimizer with initial learning rate 0.0005. We adopt plateau learning rate scheduler that reduces the learning rate by 10\% until 0.0001 when the PSNR of validation stops increasing, with the patience set to 10 for Rain100L and 3 for other datasets due to the intention of saving training time. For data augmentation, we use random horizontal flip with the possibility set to 0.5. We terminate the training when it reaches minimal learning rate (i.e. 0.0001 for all our experiments) and its validation performance stabilizes (i.e. reaches convergence). Using 8 GPUs, it takes about 2 hours to train Rain100L and 1\~{}3 days for other datasets.

\textbf{Testing.} During the test time, we keep the original input size unchanged. We validate learning outcomes after each learning epoch and we report the mean value of best validation metrics as our results.

\subsection{Results and Comparisons}
We report both quantitative (i.e. metrics) and qualitative (i.e. visualization) results on four synthetic benchmark datasets and qualitative analysis on real-world de-raining outcomes as well. We compare those results with six state-of-the-art single image de-raining methods, which are Gaussian mixture model based rain-streak removal~\cite{li2016rain}, denoted as GMM, discriminative sparse coding based de-raining strategy (DSC)~\cite{luo2015removing}, deep detail network (DDN)~\cite{fu2017removing}, deep joint rain detection and removal network (JORDER)~\cite{yang2017deep}, density-aware image de-raining using multi-stream dense network (DID-MDN)~\cite{zhang2018density} and non-locally enhanced encoder-decoder network (NLEDN)~\cite{li2018non}.

\textbf{Synthetic Datasets.} For quantitative results and comparison on four benchmark datasets, as can be seen in Table.~\ref{tab:comparison}, our proposed GradNet consistently and considerably outperforms other state-of-the-art single image de-raining methods by 2.72dB, 0.52dB, 0.98dB, 2.78dB performance gains w.r.t. PSNR and 3.16\%, 0.92\%, 0.59\%, 3.51\% gains w.r.t. SSIM, respectively on DDN-Dataset, DID-Dataset, Rain100L and Rain100H when compared with NLEDN~\cite{li2018non}, which was previously the best performer. We also find that our coarse-to-fine de-raining strategy especially effective when faced with rain streaks with high density. We believe it is because the fine de-raining stage of our method is capable of more precise image reconstruction. 

We also report the detailed boxtplot showing the distribution of our de-raining results in Figure~\ref{fig:boxplot} in Appendix, for your reference. These boxplot.pdfs illustrates that the vast majority of our testing results are closely and stably distributed, indicating the strong adaptability and prediction consistency our model possesses. We also demonstrate intermediate results of our model in Table~\ref{tab:reb-table} in Appendix, showing that our model successfully captures raining mask signals and the de-raining quality improves gradually per stage in accordance with our design and expectation. When it comes to the qualitative comparison, Fig.~\ref{fig:synthetic-qualitative} demonstrates the several de-rained samples by different methods with original inputs on the left and ground truth on the right. Through pixel-peeping the words and letters on the stone monument of the 1st example image, the steel structural support of the bridge of the 2nd example, the outdoor corridor of that white building in the 3rd sample, the girl's hair and pattern on the her clothes of the 4th example and the details of the man's umbrella along with the background environment in the 5th raining example, clear differences w.r.t. the completeness of the rain-streak removal and the quality of preserved details and reconstructed areas can be observed between our proposed GraNet and other state-of-the-art methods. These reinforced details in our results demonstrate the advantage of having a fine de-raining process. As is illustrated in Fig.~\ref{fig:synthetic interpretability}, our model successfully learns the rain-streak mask and the interpolation during the fine de-raining stage and we can see the gradually improvement of rain-streak removal and image reconstruction. By analyzing visualization results, it can be concluded that our proposed method has the best de-raining capability while preserving most details on those benchmark datasets.

\textbf{Real-world Data.} We test our proposed method (GraNet) on above-mentioned real-world images. Due to the fact that there are no ground truth for real-world raining images, we carry out qualitative analysis on the de-raining results in comparison with other state-of-the-art methods. From Fig.~\ref{fig:framework}, it is clear that our proposed method (GraNet) performs well on the real data as it learns the complicated rainy mask in reality and gradually de-rain the real-world sample. As is depicted in the Fig.~\ref{fig:real-qualitative}, we demonstrate that our model can effectively tackle with sophisticated real-world scenarios with various shapes of heavy or light rain streaks together in one rainy sample, indicating the strong generalization ability and versatility of our model, whereas many other methods can remove only the water mist or specific types of rain streaks and do not generalize well for real raining data.

\subsection{Ablation Study}
We perform the ablation study on both Rain100L and Rain100H, which are two datasets having distinct difference w.r.t. the rain-streak shape and raining density, to better illustrate the effectiveness of our model design.
\begin{table}[]
	\tabcolsep 0.15in{\scriptsize{}}
	\begin{tabular}{c|c|c}
		\hline
		\multicolumn{3}{c}{Rain100L} \\
		\hline
		Settings                    & PSNR  & SSIM   \\ \hline
		Only Coarse w/o RA   & 35.57 & 0.9693 \\
		Only Coarse w/ RA    & 36.12 & 0.9714 \\
		Coarse + Fine (w/o Context) & 35.93 & 0.9711 \\
		Coarse + Fine (w/ Context)  & \textbf{37.55} & \textbf{0.9806} \\\hline
		\hline
		\multicolumn{3}{c}{Rain100H} \\ \hline
		Settings                    & PSNR  & SSIM   \\ \hline
		Only Coarse w/o RA   & 30.45 & 0.8922 \\
		Only Coarse w/ RA    & 30.77 & 0.8964 \\
		Coarse + Fine (w/o Context) & 30.73 & 0.8979 \\
		Coarse + Fine (w/ Context)  & \textbf{33.16} & \textbf{0.9290} \\
		\hline
	\end{tabular}
	\smallskip
	\caption{Ablation Study on Rain100L and Rain100H. Rain100L mainly contains rainy images with light rain streaks. Rain100H consists of heavy rainy images.}
	\label{ablation}
	\vspace{-6mm}
\end{table}

We use the following four settings for the ablation study process. We first test the performance using only the coarse stage of our model with region-aware mechanism removed (i.e. taking away all RA Blocks), denoted as "Only Coarse w/o RA", followed by adding RA Blocks back into the first stage and still excluding the fine de-raining stage to formulate our second entry of ablation study cases, denoted as "Only Coarse w/ RA". We then develop the previous model by adding the fine de-raining stage of our proposed network but without unified contextual merging mechanism (i.e. removing dense blocks and the merging block), referred as "Coarse + Fine (w/o Context)", to test the effect of unified context modeling. At last, we provide the quantitative result of the full model entry, named "Coarse + Fine (w/ Context)", which is just the GradNet we proposed, in order to compare with all previous cases.

Results from Table~\ref{ablation} show that region-aware blocks, which are designed to enlarge receptive field, and the context fusion blocks, which are designed for de-raining under contextualized representation, are effective and useful structure of our proposed GradNet under universal cases (i.e. both heavy and light rain-streak removal task). Also, we demonstrate that without context fusion, the second stage of our proposed network will not be able to improve the overall performance of our model.

\subsection{Interpretability}
\begin{figure}[t]
	\centering\includegraphics[width=0.95\linewidth]{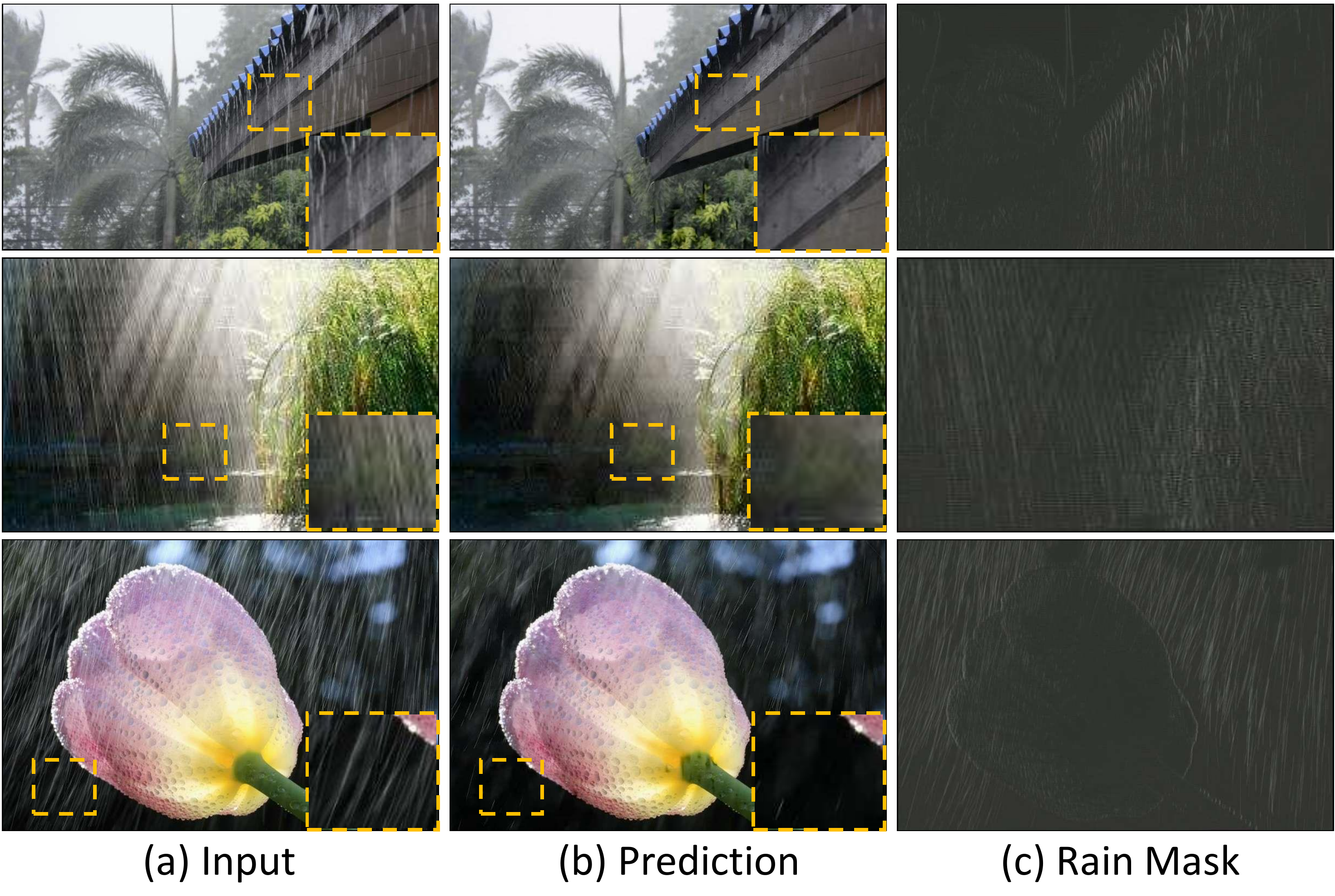}
%		\vspace{-3mm}
	\caption{Visualization results of our GraNet on real-world images. Our GraNet can generalize the capability of single image de-raining to real-world data (comparison between (a) and (b)), and can well filter the coarse-grained rain streaks as shown in (c). Please zoom in the colored PDF version of this paper for more details.}
	\label{fig:real-image interpretability}
		\vspace{-3mm}
\end{figure}

\begin{figure}[t]
	\centering\includegraphics[width=1.0\linewidth]{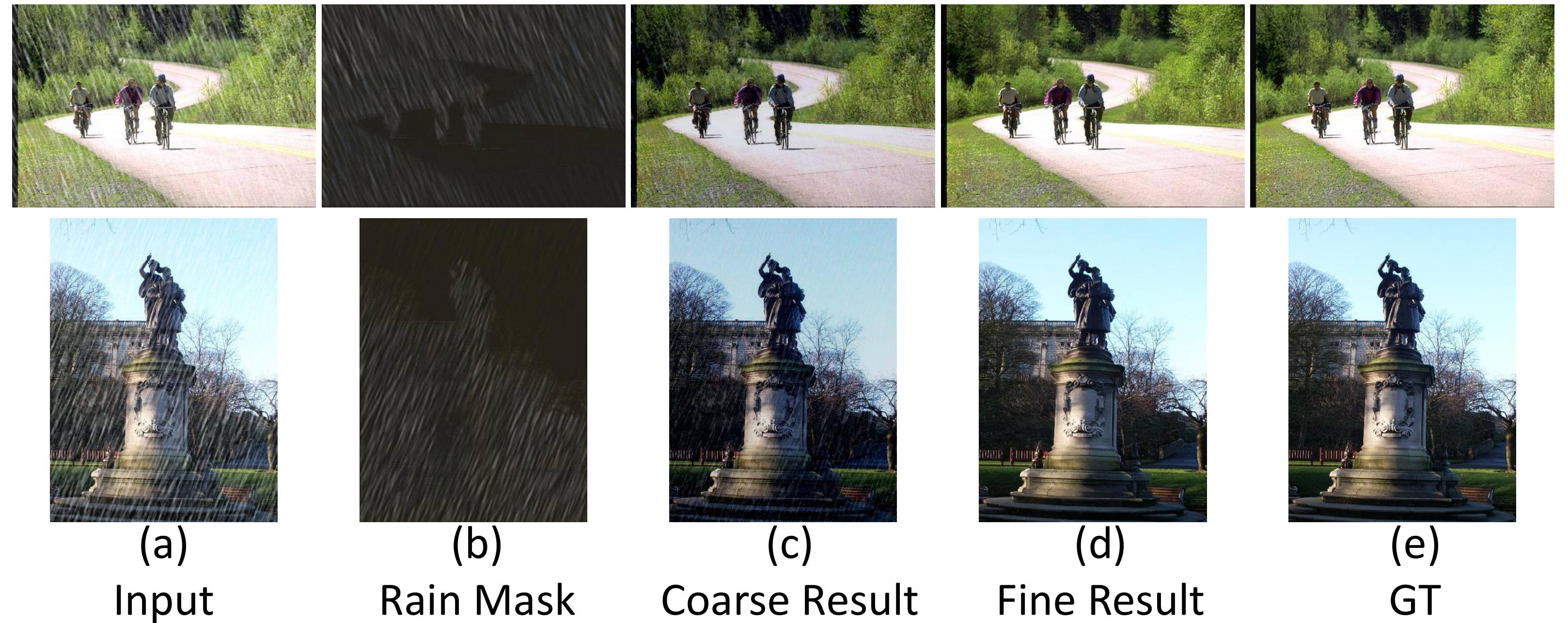}
%		\vspace{-3mm}
	\caption{The visualization of intermediate results and final result of our GraNet. (a) Input rainy images. (b) Rain mask can be also called coarse-grained rain streaks. (c) The coarse de-rained results without thick rain streaks/raindrops. (d) The fine-grained de-rained results. (e) Ground truth. These visualization can further demonstrate the effectiveness of our GraNet as we can explicitly interpret the modules and their functionalities in our GraNet. Please zoom in the colored PDF version of this paper for more details.}
	\label{fig:synthetic interpretability}
	\vspace{-4mm}
\end{figure}

We now move on to the interpretation of our results, which is another advantage that our proposed method possesses. As is shown in Fig.~\ref{fig:real-image interpretability} \& Fig.~\ref{fig:synthetic interpretability}, by visualizing the intermediate output of our model, we can clearly see what GraNet has learnt during the first stage of coarse de-raining as well as what the input of the second stage looks like, which will be used in the fine de-raining process. Both real-world examples in Fig.~\ref{fig:real-image interpretability} and artificial synthetic data in Fig.~\ref{fig:synthetic interpretability} show that by using local-global joint coarse de-raining stage, easy-separable rain-streak masks are removed by our proposed model and then by a context-aware fine de-raining process, the lost details are recovered and unhandled rain streaks disappear. Take the second sample in Fig.~\ref{fig:synthetic interpretability} (i.e. the statue image) as an instance, after the first stage of de-raining, the output mask and the coarse result (shown at the second line of Fig.~\ref{fig:synthetic interpretability}(c) \& Fig.~\ref{fig:synthetic interpretability}(d)) indicate the fact that via coarse de-raining, rain streaks that exist in the environment circling around the statue are removed as this task is relatively easier than differentiating between rain streaks and the sky which has the similar color. However, this issue is tackled with by the fine de-reaining as it utilizes contextual information and our final result (shown at the second line of Fig.~\ref{fig:synthetic interpretability}(d)) shows no or little remaining rain streaks in the picture. Through detailed visualized illustration throughout the de-raining process, we can reasonably interpret our results with concrete facts to show the effectiveness of our design.

\section{Conclusion}
In this paper, we introduced a coarse-to-fine network called Gradual Network (GraNet) for single image de-raining with different granularities, which consists of coarse stage and fine stage for mining the balance between over-de-raining and under-de-raining. The coarse stage is to deal with coarse-grained rain-streak characteristics by utilizing local-global spatial dependencies via a local-global sub-network composed of region-aware blocks. The fine stage is designed to remove the fine-grained rain streaks from coarse de-rained result and is capable of getting a rain-free and well-reconstructed outcome via a unified contextual merging sub-network composed of dense blocks and a merging block. Solid and comprehensive experiments on synthetic datasets and real data demonstrate that our GraNet can significantly outperform other state-of-the-art methods.

\section{Acknowledgements}
This work was supported in part by National Natural Science Foundation of China (NSFC) under Grant No.1811461, and in part by Natural Science Foundation of Guangdong Province, China under Grant No.2018B030312002.
We also greatly appreciate our colleagues Huabin Zheng for technical support, Yudian Li for additional computational resources and Xiaohan Zhang for last-minute proofreading.
\FloatBarrier
\clearpage
\bibliographystyle{ACM-Reference-Format}
\bibliography{acmart}
\FloatBarrier
\clearpage
% \pagebreak
% \FloatBarrier
% \pagebreak
\onecolumn
\appendix
\section{Appendices}
% \pagebreak
%\newpage
% \FloatBarrier
% \vspace{-100mm}
% \FloatBarrier
\begin{figure*}[!htb]
	% \vspace{30mm}
	\centering\includegraphics[width=0.9\linewidth]{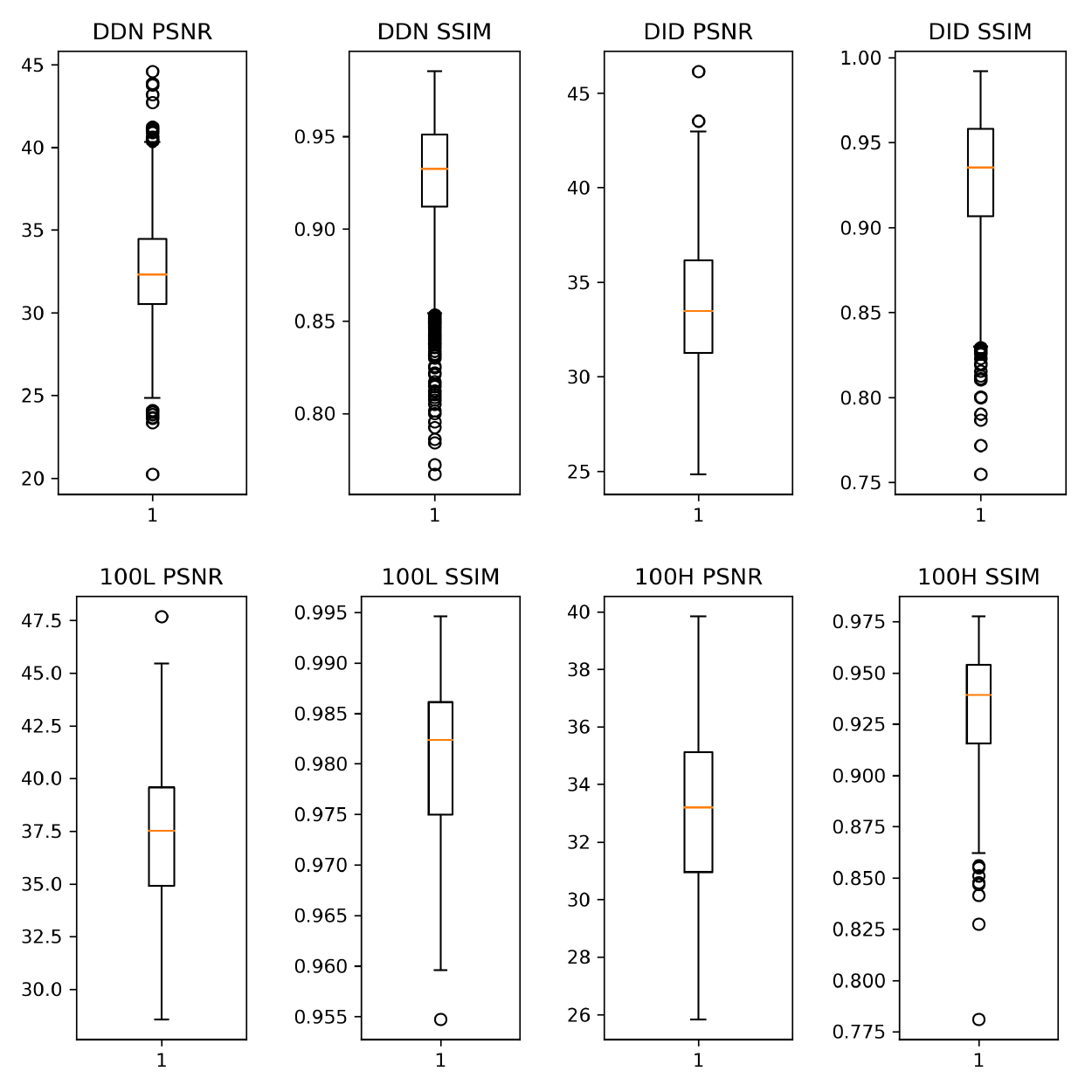}
	\vspace{-6mm}
	\caption{Quantitative boxplot results of PSNR \& SSIM of our model on four synthetic datasets. The {\color{red}{red}} line indicates the median with the box showing the range from lower to upper quartile. Flier points are considered outliers.}
	\label{fig:boxplot}
	% 	\vspace{-3mm}
\end{figure*}
\begin{table*}[!htb]
	\begin{tabular}{cc|cccc}
		\hline
		&  & DDN-Dataset & DID-Dataset & Rain 100L & Rain 100H \\ \hline
		\multirow{2}{*}{Coarse} & PSNR & 19.66 & 19.56 & 23.10 & 18.02 \\
		& SSIM & 0.7389 & 0.7214 & 0.7947 & 0.4791 \\ \hline
		\multirow{2}{*}{Mask} & PSNR & 19.81 & 19.48 & 23.50 & 17.86 \\
		& SSIM & 0.3374 & 0.3650 & 0.1442 & 0.5531 \\\hline
	\end{tabular} 
	\vspace{2mm}
	\caption{Quantitative results of PSNR \& SSIM of the coarse output (i.e. the de-rained outcome after the coarse stage) and the mask output (i.e. the predicted rain mask from the first stage).}
	\label{tab:reb-table}
\end{table*}

\end{document}